\documentclass[conference]{IEEEtran}
\IEEEoverridecommandlockouts
\usepackage{cite}
\usepackage{amsmath,amssymb,amsfonts}
\usepackage{algorithmic}
\usepackage{graphicx}
\usepackage{textcomp}
\usepackage{xcolor}
\usepackage{makecell}
\usepackage{graphicx,booktabs,latexsym}
\usepackage[lined,boxed,ruled]{algorithm2e}
\usepackage{subfigure}
\usepackage[switch]{lineno}
\usepackage{hhline}
\usepackage{multirow}
\usepackage{multicol}
\usepackage{listings} 
\usepackage[numbers]{natbib}
\usepackage{hyperref}

\def\BibTeX{{\rm B\kern-.05em{\sc i\kern-.025em b}\kern-.08em
    T\kern-.1667em\lower.7ex\hbox{E}\kern-.125emX}}
\begin{document}

\title{NeuronsMAE: A Novel Multi-Agent Reinforcement Learning Environment for Cooperative and Competitive Multi-Robot Tasks 
}
\author{
Guangzheng~Hu,~\IEEEmembership{Student Member,~IEEE},
Haoran~Li,~\IEEEmembership{Member,~IEEE},
Shasha~Liu,~\IEEEmembership{Student Member,~IEEE},\\
Mingjun~Ma,~\IEEEmembership{Student Member,~IEEE},
Yuanheng~Zhu,~\IEEEmembership{Senior Member,~IEEE},
and~Dongbin~Zhao,~\IEEEmembership{Fellow,~IEEE}

\thanks{G. Hu, S. liu, and M. Ma are with the School of Artificial Intelligence, University of Chinese Academy of Sciences, Beijing 100049, China, and also with the State Key Laboratory of Multimodal Artificial Intelligence Systems, Institute of Automation, Chinese Academy of Sciences, Beijing 100190, China (email : \{huguangzheng2019, , liushasha2020, mingjun.ma\}@ia.ac.cn).}

\thanks{Y. Zhu, H. Li and D. Zhao are with the State Key Laboratory of Multimodal Artificial Intelligence Systems, Institute of Automation, Chinese Academy of Sciences, Beijing 100190, China, and also with the School of Artificial Intelligence, University of Chinese Academy of Sciences, Beijing 100049, China (email : \{yuanheng.zhu, lihaoran2015, dongbin.zhao\}@ia.ac.cn).}
}

\maketitle

\begin{abstract}
Multi-agent reinforcement learning (MARL) has achieved remarkable success in various challenging problems. Meanwhile, more and more benchmarks have emerged and provided some standards to evaluate the algorithms in different fields. On the one hand, the virtual MARL environments lack knowledge of real-world tasks and actuator abilities, and on the other hand, the current task-specified multi-robot platform has poor support for the generality of multi-agent reinforcement learning algorithms and lacks support for transferring from simulation to the real environment. Bridging the gap between the virtual MARL environments and the real multi-robot platform becomes the key to promoting the practicability of MARL algorithms. This paper proposes a novel MARL environment for real multi-robot tasks named NeuronsMAE (Neurons Multi-Agent Environment). This environment supports cooperative and competitive multi-robot tasks and is configured with rich parameter interfaces to study the multi-agent policy transfer from simulation to reality. With this platform, we evaluate various popular MARL algorithms and build a new MARL benchmark for multi-robot tasks. We hope that this platform will facilitate the research and application of MARL algorithms for real robot tasks. Information about the benchmark and the open-source code will be released.
\end{abstract}


\section{Introduction}\label{sec:introduction}

%
%
%
%

Tracing back to the past few years since deep Q-learning, deep reinforcement learning (DRL) has been playing a significant role and applied to solve challenging problems from the game \cite{shao2018starcraft} to robotics \cite{dexterous}. As an extension, MARL, carried by the advance of single-agent DRL, has seen revolutionary breakthroughs with its successful application to many scenarios where multiple agents need to cooperate to accomplish complicated tasks \cite{masurvey}, and a plethora of literature has emerged lately \cite{vdn}\cite{qmix}\cite{mappo}\cite{happo}.

Due to multi-agent characteristics such as non-stability and limited communication, MARL is still a very challenging topic. In order to fairly evaluate the performance of different algorithms, various environments and benchmarks have emerged. Perhaps inspired by the landmark breakthrough of deep reinforcement learning in video games, the game platforms have become one of the choices for developing and evaluating MARL algorithms \cite{smac, wei2022honor, grf, jaderberg2019human}. These platforms build multi-agent tasks by leveraging the game's natural multi-player collaboration or competition properties. Others are evolved from the single-agent task, which decomposes the single-agent control task into multi-agent tasks, such as MaMujoco \cite{mamujoco}, and SMARobosuite \cite{gu2021multi}. In addition, there are also some environments \cite{mpe, mlagents, magent} that simplify real-world tasks by abstracting them or completely imagining virtual tasks for building collaborative or competitive multi-agent tasks. Most of these platforms and tasks are currently limited to simulation environments, which do not take into account features such as real-world task characteristics and the execution limitations of the real actuators, and are therefore difficult to use for the research and application of MARL algorithms in real multi-robot tasks.

Multi-robot systems also have a long history of development, but the current multi-robot platform is still dominated by collaborative tasks, such as multi-arm cooperation platforms \cite{tung2021learning, ha2021learning}, dexterous hand cooperation platforms \cite{dexterous}, multi-mobile robot cooperation platforms \cite{chen2022multirobolearn, liang2022multi}, and so on. There are also some multi-agent platforms for competitive tasks, such as the RoboCup-derived soccer platform Bassani 2020 Framework. In general, the current multi-robot platform is usually task-specific. There are few widely used multi-robot research and application platforms due to the task dependence on the robot platform and the diversity of actuators. Moreover, most current platforms only consider the task characteristics of multi-robot. They rarely consider the needs and challenges faced by the multi-agent policy transferring from the simulation to the real-world environment. Although there is a wealth of research on single-agent tasks, it is still a worthwhile direction to explore for multi-agent policy transfer.

Because there is a gap between the virtual MARL benchmark and the real multi-robot application, figuring out how to build a universal platform that takes both MARL characteristics and multi-robot application attributes into account has become critical to resolving the quandary. Considering the cooperative and competitive attributes of multi-agent tasks, the practical application platform of multi-robots, and potential beneficiaries, we propose a novel multi-robot policy training and evaluation platform for cooperative and competitive tasks, NeuronsMAE, based on the currently popular RoboMaster University AI Challenge. Specifically, our main contributions include the following:
\begin{itemize}
	\item We propose a multi-agent algorithm training and evaluation platform for multi-robot tasks named NeuronsMAE. This platform aims at cooperative and competitive tasks for robots in the real world, including highly flexible observation and action spaces. It can customize the state observability and action attributes according to the algorithm's characteristics.
	
	\item NeuronsMAE provides a high-fidelity environment and robot model, as well as rich parameter interfaces, to support the research of multi-robot policy transfer from a simulation to a real-world environment.
	
	\item We evaluate several commonly used MARL algorithms on NeuronsMAE and create a new benchmark for multi-robot cooperative and competitive tasks in order to promote MARL algorithm research and application in the field of multi-robot. 
\end{itemize}

\section{Related Work}\label{sec:relatedWork}

This section will review the current mainstream MARL benchmark and multi-robot platform, respectively.

\subsection{MARL Benchmarks}

Multifarious emerging benchmarks have accelerated MARL research in recent years and provide various evaluation criteria for different application scenarios and research domains. Some game platforms have been the most popular benchmarks for evaluating MARL algorithms. For example, SMAC \cite{smac} is based on the popular real-time strategy game StarCraft II, focuses on micromanagement challenges, and is applicable to studying cooperative MARL. GRF \cite{grf}, an environment for playing football tasks of varying difficulty in a physics-based 3D simulation, focuses on multi-level, multi-agent learning. Wimblepong \cite{zhu2022empirical} is a 2-player version of the Atari game Pong, where each player controls a paddle to play a ball with the other, and it is a purely competitive scenario. Some environments evolve from single-agent tasks, which decompose single-agent control tasks into multi-agent tasks. For instance, MaMujoco \cite{mamujoco},  based on a single-agent robotic MuJoCo control suite, provides a wide variety of continuous multi-agent robotic control scenarios in which multiple agents within a single robot try to complete a task cooperatively. DM\_Control \cite{dmcontrol} is a set of Python RL environments powered by the MuJoCo physics engine and includes multi-agent soccer simulation environments. Others are environments that abstract real-world tasks or completely imagine virtual tasks in order to build multi-agent tasks. MPE \cite{mpe} is a multi-agent particle world environment with several cooperative and competitive scenarios. MAgent \cite{magent} focuses on supporting competitive tasks that require hundreds to millions of agents. Unity Machine Learning Agents Toolkit (ML-agents) \cite{mlagents} is an open-source project with 17+ example environments for training intelligent agents. These abovementioned platforms and environments are currently limited to simulation environments, which do not take into account features such as real-world task characteristics and the execution limitations of the real actuators. Consequently, it is difficult to use them for the research and application of MARL algorithms in real multi-robot tasks.

\subsection{Multi-Robot Platforms}

There are many research work and platforms for multi-robot systems, the most common of which is the multi-manipulator cooperation task. For example, the multi-arm cooperation platform \cite{tung2021learning} proposes a variety of two-arm cooperation platforms to evaluate the performance of multi-agent cooperation policies on multi-arm cooperation tasks. Kennel-Maushart et al. \cite{kennel2021manipulability} define the task for multi-arm cooperation in surgical robots. Ha et al. \cite{ha2021learning} define different levels of tasks for the planning of three-arm cooperation, which is used to evaluate the capabilities of multi-agent policies. In addition, Gu et al. \cite{ha2021learning} adopt a similar method to that in MaMujoco, defining the controllers of different joints on a single manipulator as different agents, thus converting the control of single or multiple robot arms to a multi-agent control problem, and design a variety of tasks for the cooperation of two dexterous hands. In addition to manipulator tasks, multi-mobile robots are a commonly used platform. Cui et al. \cite{cui2022learning} design the task of interactive navigation of multiple mobile robots with partial observation based on the TurtleBot. In addition to the above collaborative tasks, there are also some competitive multi-robot systems. For example, the VSSS-RL framework proposed for the IEEE VSSS Challenge is committed to applying multi-agent deep reinforcement learning algorithms to small soccer systems \cite{bassani2020framework}. The RoboMaster series, represented by RoboMaster University Challenge (RMUC), RoboMaster University AI Challenge (RMUA), and RoboMaster Young Challenge (RMYC), also focus on competitive robot tasks, but mainly involve classic robot technologies such as mechanical design, embedded development, etc. The NeuronsMAE is based on the RMUA, which simplifies the task setting and hardware requirements and makes the research focus more on the algorithm and multi-robot task itself, which has a broader application significance. On the other hand, current work in multi-robot systems focuses primarily on the task, ignoring the process of policy transfer from simulation to a real-world environment. Although this part has been widely studied in the single-agent field, the impact of the Sim2Real problem on multi-agent policy transfer cannot be accurately assessed. Our NeuronsMAE provides rich parameter interfaces that can be used to support multi-agent policy transfer research.

\section{Neurons Multi-Agent Environment}\label{sec:benchmark}

We built the NeuronsMAE platform on RMUA, taking into account the cooperative and competitive characteristics of multi-agent tasks, the universality of robot platforms, and potential beneficiaries. According to the rules of the RMUA, robots need to cooperate with teammates and fight opponents by shooting. The robot needs to defeat the opponents as much as possible within a certain time limit to win the game.

\begin{figure*}[!t]
	\centering
	\begin{center}
		\scriptsize
		\includegraphics*[width=5.5in]{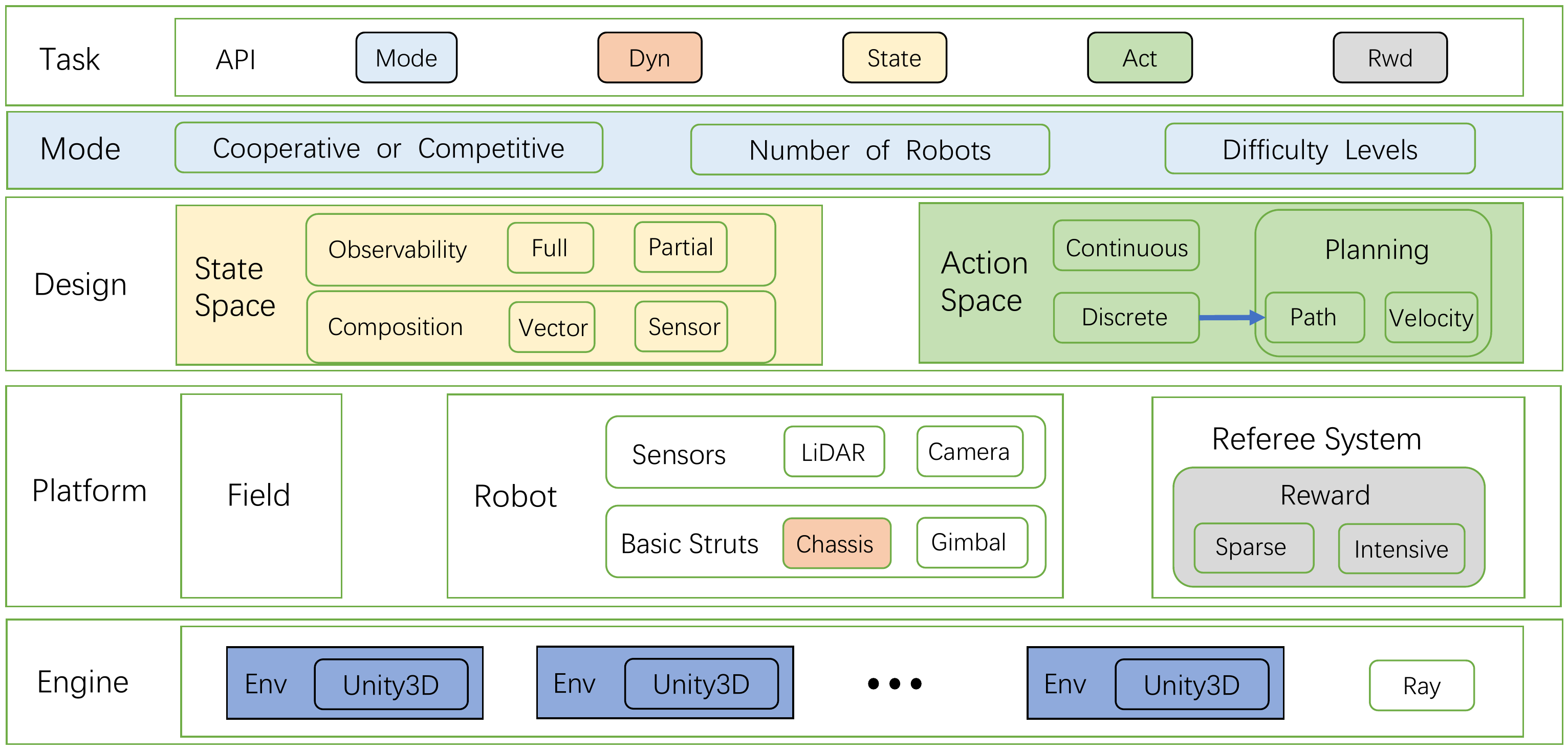}\\
		\caption{The structure diagram of NeuronsMAE. The bottom layer is the engine, based on which all functions can be implemented. The platform layer consists of a field, robot, and referee system. In the upper layer, we design several different state spaces and action spaces. The mode layer can set the task characteristics and the number of robots. The top layer consists of the API and parallelization. API interfaces allow users to configure the environment for various scientific problems. Each interface is connected to the module in the lower layers. The corresponding relationship between them can be seen from the color matching in the diagram.}
		\label{fig:struts}
	\end{center}
	\vspace{-0.5cm} 
\end{figure*}

NeuronsMAE is developed based on the Unity3D engine. We build a platform for multi-robot cooperation and competition, develop multiple state spaces and action spaces, design various modes with diverse characteristics, and provide the MARL-fitted tasks to cover as many research domains as possible. The structure diagram of NeuronsMAE, as shown in Figure \ref{fig:struts}, presents the hierarchic dividing of system construction, which consists of multiple modules to complete specific functions. In this section, we detail NeuronsMAE and its multiple modules to show its advantages of being applied to various research domains in MARL.

\subsection{Engine}\label{sec: engine}

Physical engines play an essential role in all environments. As with current MARL benchmarks, trade-offs must be weighed between processing speed, fidelity, and other factors. Unity3D has a high degree of fidelity in its physical engine and allows various materials and sensors to be added to the environment. Above all, the environment can achieve speeds of up to 100x, significantly reducing the time required to interact with it. Therefore, we adopt Unity3D as our physical engine, in which all entities are created as close to the real world as possible.

In addition to the physical engine of the simulation, we utilize a high-performance distributed computing engine named Ray. We can obtain enough data for training as soon as possible within an acceptable time by seamlessly scaling the sampling process. Ray is a unified framework for increasingly compute-intensive tasks and scaling such applications \cite{ray}. One can set the number of parallel sampling processes according to the hardware conditions and the number of waiting batch sampling, based on which synchrony and asynchrony can be indirectly set.

\subsection{Platform}\label{sec:platform}

\begin{figure*}[!t]
	\centering
	\begin{center}
		\scriptsize
		\includegraphics*[width=5.0in, height=2.7in]{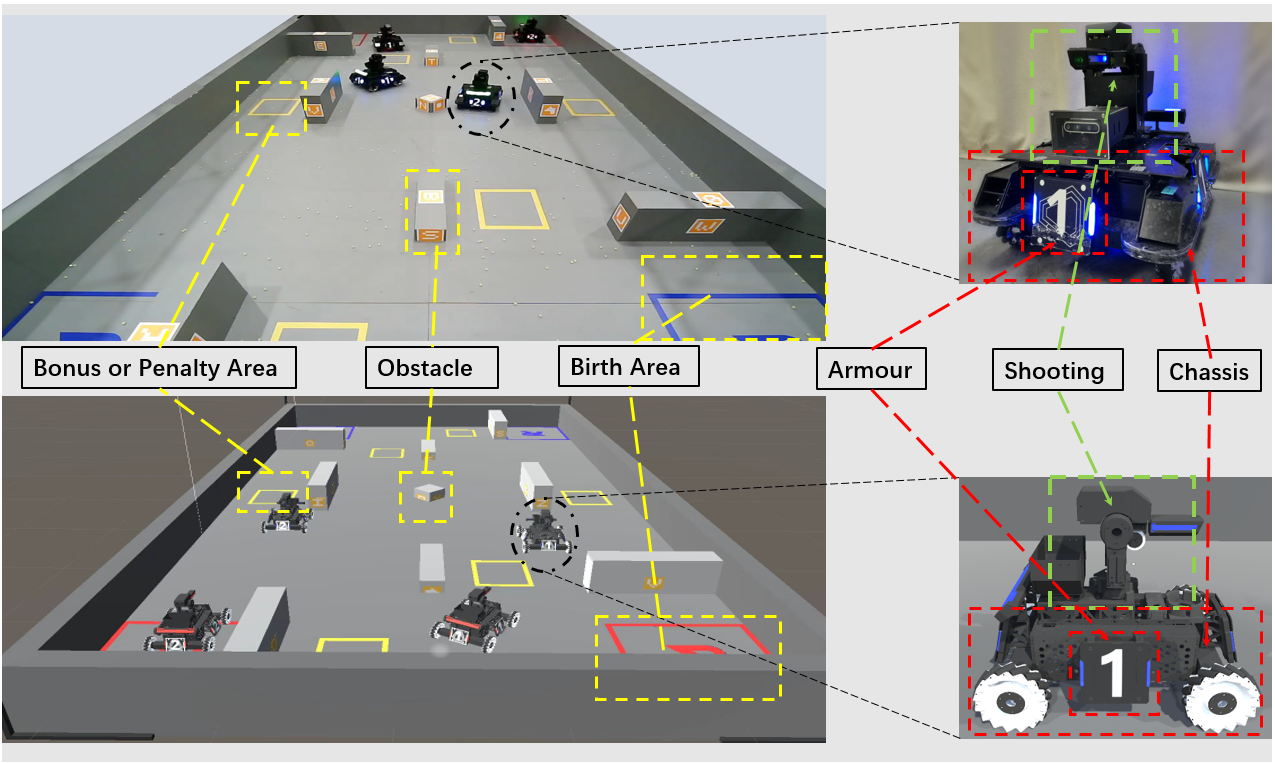}\\
		\caption{Illustration of the platforms in the real world and NeuronsMAE. At the top is the platform in the real world, and below that is the visualization of the platform in NeuronsMAE. The arena in NeuronsMAE is the same as the real world, including birth areas, bonus areas, and penalty areas. We can see each obstacle's length, width, and height from the two perspectives. In addition, the full view of a robot and its components is presented on the right.}
		\label{fig:platform}
	\end{center}
	\vspace{-0.5cm} 
\end{figure*}

Based on Unity3D, we have built a platform for multi-robot cooperative competition. As shown in Figure \ref{fig:struts}, the platform consists of three parts: the field, the robot, and the referee system. The field part introduces the composition of the whole scene, the robot part details the components of the robot, and the referee system part presents the specific functions of the referee system. The general overviews of the platforms in NeuronsMAE and the real world are presented in Figure \ref{fig:platform}. Next, we will introduce each part in detail.

\noindent\textbf{Field }
The competition arena is 8.1 meters long and 5.1 meters wide. Four birth areas located at the four corners are used as the initial positions of the robots. In addition, six bonus or penalty areas and nine obstacles of different heights and shapes have cylindrical symmetry in polarization.

\noindent\textbf{Robot }
A robot is composed of sensors such as LiDAR and a camera, and the basic structure includes a chassis module and a gimbal module.

The chassis module of the robot has four Mecanum wheels, which allow the robot to move in any direction.
The chassis is equipped with four pieces of armor around it. The attack is only effective if the bullets hit the armor.

The robot shooting module consists of the gimbal submodule and the shooting submodule.
The camera and the shooting submodule are mounted on the gimbal submodule, which can complete the rotation movement of two degrees of freedom: pitch and yaw. The gimbal improves the flexibility and competitive ability of the robot. The shooting submodule can launch bullets to hit the opponent's armor.

In the real-world physical environment, the trajectory of the projectile is related to the wear degree of the launching mechanism, the projectile speed, the moving speed of the robot, the air density and humidity, and many other conditions. In that context, it is very complex to conduct accurate calculations in the simulation, and it is difficult to meet the hit rate of the real-world environment. Therefore, we have not realized the control of the shooting module to launch bullets in the simulation. Instead, we counted the hit rate at different shooting distances and fitted the function between hit rate and distance. By means of the function, the referee system can easily calculate the HP loss of each robot, the expectation of which is the same as the real-world one.

\noindent\textbf{Referee System }
The main functions of the referee system include:
\begin{itemize}
	\item Information summary: summarize the information on both sides, such as HP, bullet quantity, and so on.
    \item HP accounting. If a robot's armor is hit, the referee system will judge how much HP should be deducted from the hitting probability and which armor.

	\item Bullet accounting. If a robot executes the shooting command, the referee system will deduct a certain number of remaining bullets.
	
	\item Judgment. The following are the criteria for winning in a single round: (1) The match ends immediately when all the robots of a team are destroyed, and the team with the surviving robot(s) wins; (2) when the entire time of a round is up, if robots from both teams have survived, the team with the higher damage output wins.
\end{itemize}

The overall goal of each team is to maximize its win rate. The referee system in NeuronsMAE provides three optional out-of-the-box reward functions: hit-point damage dealt, enemy units killed, and a bonus for winning the battle. Users can use only rewards for winning or losing an episode when studying sparse reward problems, or they can utilize all reward functions as a series of intensive rewards in order to achieve better performance. It also allows researchers to add custom reward functions using wrappers, for example, a penalty for a collision with obstacles or other robots.

\subsection{State Space}\label{sec: statespace}
We create vector-based and sensor-based state spaces to aid algorithmic research on MDP in multi-mode states.

\begin{figure}[!t]
	\centering
	\begin{center}
		\scriptsize
		\includegraphics*[width=2.0in, height=1in]{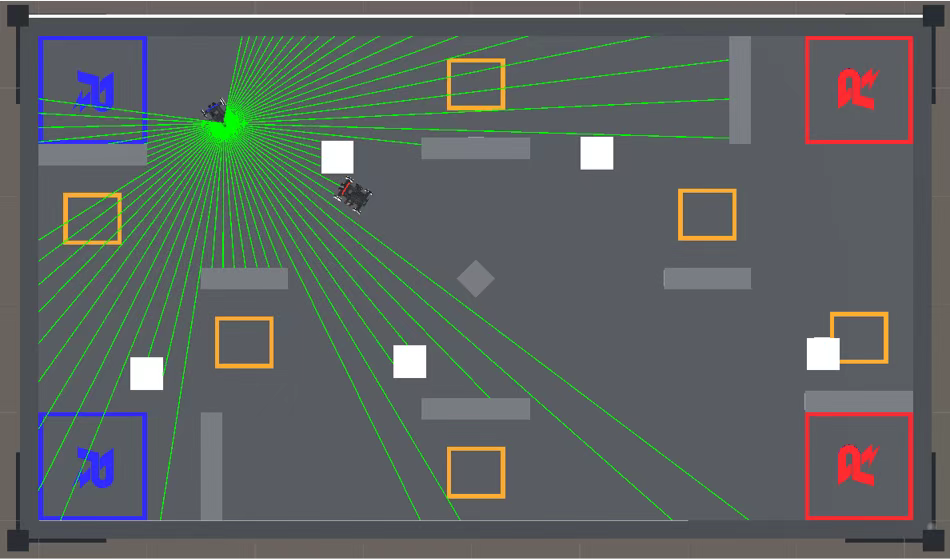}\\
		\caption{The sensor effect in NeuronsMAE. The LiDAR can detect the obstacle of $270^{\circ}$  ahead.}
		\label{fig:sensor}
	\end{center}
\vspace{-0.5cm} 
\end{figure}

Vector-based states are obtained from the referee system and can either be extended to include all information or not for the studies on the fully or partially observable MDPs. As shown in Table \ref{tab:vector}, the vector-based state information is composed of the pose, candidate position (detailed in Section \ref{sec:demo}), HP, and bullet count of all or some robots. The sensor-based information stems from LiDAR, as shown in Figure \ref{fig:sensor}.

All of those, as mentioned above, are optional for users and could be arbitrarily combined for different requirements. The following example is representative to illustrate how to use it. We suppose a robot can obtain information about itself and its ally, HP, and bullet count from the referee system. However, one robot can calculate the positions and angles of opponents only when its camera captures them, or pad the unobserved information with a specific value.

\begin{table}[t]
	\centering
	\caption{{The complete state information in NeuronsMAE.}}
	\begin{tabular}{|c|c|c|}
		\hline \hline
		{\bf Attribute} &{\makecell[c]{\bf Dimension of \\ \bf each object}} & {\bf Object}   \\ \hline
		{Pose} & {3} & {\makecell[c]{All robots}}   \\ \hline
		{Position} & {2} & {\makecell[c]{All candidate points}}   \\ \hline
		{HP} & {1} & {All robots}   \\ \hline
		{Bullet} & {1} & {All robots}   \\ \hline
		{Time} & {1} & {Referee System} \\ 
		\hline \hline
	\end{tabular}%
	\label{tab:vector}%
	\vspace{-0.3cm} 
\end{table}%

\subsection{Action Space}\label{sec:actionspace}
In most of the mobile robot research, the commonly used action spaces can be divided into three groups: continuous action spaces \cite{leiva2020robustcont}, discrete action spaces \cite{yokoyama2020autonomousdis} and abstract action spaces \cite{martin2019variableabstract}. Continuous action space usually refers to the expected velocity of the robot. Similarly, the discrete action space is the discrete velocity space. Abstract action space is often used in conjunction with planners and controllers. For example, it usually contains lane-changing instructions in the field of automatic driving and uses target points in the field of indoor mobile robots.

Unlike mobile robots, the task in NeuronsMAE is composite, which controls the robot's movement and the attacking target. Therefore, the action space of NeuronsMAE will include two parts: movement and competition.

In the part of movement action space, we consider the support of existing algorithms and design the continuous action space and abstract action space. Since the discrete action space can be easily completed based on a continuous one, there is no interface for it.
The continuous action space refers to the expected horizontal velocity, longitudinal velocity, and angular velocity of the robot. The abstract action space selects a goal point, which is the point that the robot chooses to go from several candidate points, offered by a rule-based bot (detailed in Section \ref{sec:demo}). It frees users from having to control robot motion by using a path planning algorithm and a simple velocity planning algorithm.

In the competition part, we design the discrete action space for selecting the target enemy, representing which one should be hit. It is effective for shooting only when the enemy is within the range and angle. All actions are presented in Table \ref{tab:action}. One can construct various action spaces by composing different above actions. 

\begin{table}[t]
	\centering
	\caption{{The complete actions in NeuronsMAE.}}
	\begin{tabular}{|c|c|c|}
		\hline \hline
		{\bf Action} & {\bf Attribute} & {\makecell[c]{\bf Range or \\ \bf Dimension}}   \\ \hline
		{\makecell[c]{horizontal \\ velocity}} & {Continuous} & {$-2 \sim 2$ m/s}   \\ \hline
		{\makecell[c]{longitudinal \\ velocity}} & {Continuous} & {$-2 \sim 2$ m/s}   \\ \hline
		{\makecell[c]{angular \\ velocity}} & {Continuous} & {$-1.75 \sim 1.75$ rad/s}   \\ \hline
		{\makecell[c]{goal point}} & {Abstract} & {4}   \\ \hline
		{\makecell[c]{target enemy}} & {Discrete} & {2}   \\ 
		\hline \hline
	\end{tabular}%
	\label{tab:action}%
	\vspace{-0.5cm} 
\end{table}%

\subsection{Mode}\label{sec:mode}
NeuronsMAE provides many choices for the mode by which the users can select the cooperative or competitive tasks, the number of robots, and the difficulty level.

In cooperative tasks, the environment controls the opponent team by employing a rule-based bot based on finite state machines. In competitive tasks, any trained model can be used to replace the opponent's bot.

The number of robots on both sides can vary according to the research problems. On the simplest level, one can set up an agent to compete with an opponent for single-agent RL.

The difficulty level can be set in a number of ways. Users can define the difficulty level of the task by enhancing or weakening the properties of robots, such as HP and velocity; i.e., the higher the initial HP or velocity the opponent robot has, the more difficult the task. We also design different rule-based bots with different levels of difficulty. The easy bot goes to the nearest candidate point to attack the nearest robot. The middle bot goes to the nearest candidate point to attack the robot with the least HP. The hard bot makes two robots siege the robot with the least HP at the best candidate points.

\subsection{Task}\label{sec:task}

The task layer is the API. NeuronsMAE is out-of-the-box compatible with the widely used OpenAI Gym API and almost identical to ma-gym. One can open multiple environments in different ports with different seeds to ensure the diversity of initialization.

Furthermore, API provides users with five interfaces for configuring all task characteristics based on their research domains.As shown in Figure \ref{fig:struts}, each interface is connected to the module in the lower layers. The corresponding relationship between them can be seen from the color matching in the diagram.

In addition, NeuronsMAE provides an interface for changing the dynamic parameters, as detailed in Section \ref{sec:context}. We consider randomizing the dynamic parameters in simulation in order to cover the real distribution of the real-world data despite the bias between the model and the real world.

\subsection{Interfaces for Sim2Real}\label{sec:context}
With NeuronsMAE, we provide a variety of interfaces that can influence a robot’s learning and generalization capabilities. First, an interface to support modifying the hitting probability is provided.In addition, we borrow the possible dynamic parameters that cause the variability of the state transition function from NeuronsGym \cite{haoran2023}, and describe each parameter in detail below.
\begin{itemize}
	\item Friction Coefficients: With the mobile robot dynamics analysis, the friction coefficients include two parameters: the sliding friction coefficient and the rolling friction coefficient.
	\item Motor Character: The motor character responds to the ability to generate torque, directly affecting the robot’s acceleration and braking capabilities. 
	\item Controller Parameters: We use a PID controller for the robot in the simulator and the physical robot, and the parameters are crucial to the controller.
	\item Robot Mass: The mass $M$ of the robot affects the magnitude of the inertia and, thus, the control system.
	\item Rotational Inertia of the Wheel: As can be seen from the dynamics model of the robot, the rotational inertia of the wheels affects the calculation of the velocity. 
\end{itemize}

\subsection{Baselines}\label{sec:demo}
Here, we first give an example environment for reference. The robot has several candidate target points offered by means of a rule-based bot to choose from. First, we sample 180 points with the same interval on each circle centered on an opponent and with a radius equal to the best shooting distance. And then get rid of the unreasonable points, for example, those that are inside or close to obstacles or outside the field. After that, $k$ centres can be obtained by clustering the remaining points by use of K-means \cite{likas2003global} according to the expected number of candidate points $k$, shown as the three red circles. These centers may be unreasonable, and therefore we choose the point closest to the center of each cluster. Now that $k$ candidate points can be obtained, the robot will decide which point should go there. Second, given a specific target point, the built-in bot plans a feasible path in a short time for each robot by using any traditional path planning algorithm. Then a simple velocity planning algorithm calculates the robot's velocity at each point of the path so that it can be controlled to move quickly and smoothly. When the opponent to be hit is selected, the robot will rotate in the correct direction with a reasonable angular velocity and eventually towards the opponent. To use the environment, look at the code for importing them as follows:

\lstset{
	columns=fixed,       
	numberstyle=\tiny\color{gray},                       
	frame=none,                                          
	backgroundcolor=\color[RGB]{245,245,244},            
	keywordstyle=\color[RGB]{40,40,255},                 
	numberstyle=\footnotesize\color{darkgray},           
	commentstyle=\it\color[RGB]{0,96,96},                
	stringstyle=\rmfamily\slshape\color[RGB]{128,0,0},   
	showstringspaces=false,                              
	language=python,                                        
}
\begin{lstlisting}
import RmcooEnv as RM
env = RM(seed=1234, port=10000, 
	no_graphics=True, time_scale=100)
contexts = {"VK1":0.1, "level":1}
obs = env.reset(contexts)
\end{lstlisting}
A more detailed description of the environments is given as follows:

\noindent\textbf{Observation Space and Observability}
The environment is full of observability.
The state and observation consist of information about itself, its ally, and two opponents, including pose, candidate points, HP, bullet count, and the time remaining ($ 12 + 16 + 4 + 4 + 1 = 37$-dimensional). 

\noindent\textbf{Action Space}
This environment contains abstract and discrete action spaces. 
The robot just needs to decide which point should go ($k=4$ and 4-discrete) and which opponent should be hit (2-discrete). 

\noindent\textbf{Reward}
This environment adopts three reward functions: hit-point damage dealt ($r_h = 0.02$), enemy units killed ($r_k = 3$) and a bonus for winning the battle ($r_w = 20$).

\noindent NeuronsMAE provides three MARL algorithms as follows:

\noindent\textbf{IQL} \cite{iql} extends DQN to the decentralized multi-agent reinforcement learning environment so that it can deal with a high-dimensional and complex environment.

\noindent\textbf{VDN} \cite{vdn} assume that the joint action-value function is taken as the sum of the local action-value functions.

\noindent\textbf{QMIX} \cite{qmix} assumes that the joint action-value function is monotonically decomposed into nonlinear combinations of local action-value functions.



\section {Experiments}\label{sec:Experiment}

Having discussed a set of benchmarks in NeuronsMAE and a hierarchic design and demonstration, we now use three cooperative scenarios to verify the feasibility and superiority of various algorithms in the simulation and the real world. These scenarios are presented in Section \ref{sec:task} in different difficulty levels.
We evaluate several representative MARL algorithms in the abovementioned scenarios and report baseline results in the simulation and the real world. All experiments can be reproduced using the scripts we provide with the benchmark library.

\subsection{Experimental Setup}\label{sec:setup}

The maximum length of competition time is 20.0s, and the finest grain of physical execution time (Fixed Timestep) in 0.02s. Each interaction step in the RL framework consists of 20 (alternative) Fixed Timestep and consumes 0.4s. Hence, each entire trajectory has a maximum of 50 steps.

The experiments are conducted on an Intel 72-core Xeon Gold 6140 CPU and an Nvidia Titan XP GPU. In this configuration, the environment interaction can be accelerated up to five times the original speed, i.e., each entire trajectory consumes about 4.0s. In order to obtain enough data for training as soon as possible within an acceptable time, we use Ray to seamlessly scale the sampling process from a single process to a multi-process. With Ray, we parallelize $20 \sim 25$ sampling processes at the same time in our experiments.

\subsection{Comparison in Simulation}\label{sec:simulation}
We conduct the experiments in three cooperative scenarios with different difficulty levels, compare the performances, and list the results in Table \ref{tab:results}.

\begin{table}[t]
	\centering
	\caption{Comparison among five MARL baselines in three cooperative scenarios with different difficulty levels.}
	\begin{tabular}{|c|c|c|c|}
		\hline \hline
		{\bf Level} & {\bf IQL} & {\bf VDN} & {\bf QMIX}  \\  \hline
		{Easy} & {0.75} & { 0.93} & {0.89}   \\  \hline
		{Middle} & {0.48} & { 0.62} & {0.57}   \\  \hline
		{Hard} & {0.23} & { 0.41} & {0.34}  \\ 
		\hline \hline
	\end{tabular}%
	\label{tab:results}%
	\vspace{-0.5cm} 
\end{table}%

The results and plots indicate that IQL is far inferior to other methods in all scenarios because of the non-stationary problem caused by ignoring that other agents with changing policies are part of the environment. VDN and QMIX are more suitable than IQL and give expression to the superiority of monotone decomposition. Significantly, VDN outperforms all baselines, and the performance gap between VDN and others is relatively big. Normally, QMIX is superior to VDN because of its nonlinear representation ability. However, these scenarios are fully observable, and the nonlinear representation ability of local action value function networks is enough, so the simpler network of VDN helps find the optimal solution more quickly.

\begin{figure*}[!t]
	\begin{center}
		\subfigure[{The easy scenario.}]{
			\label{fig:easy}
			\includegraphics[width=2.0in]{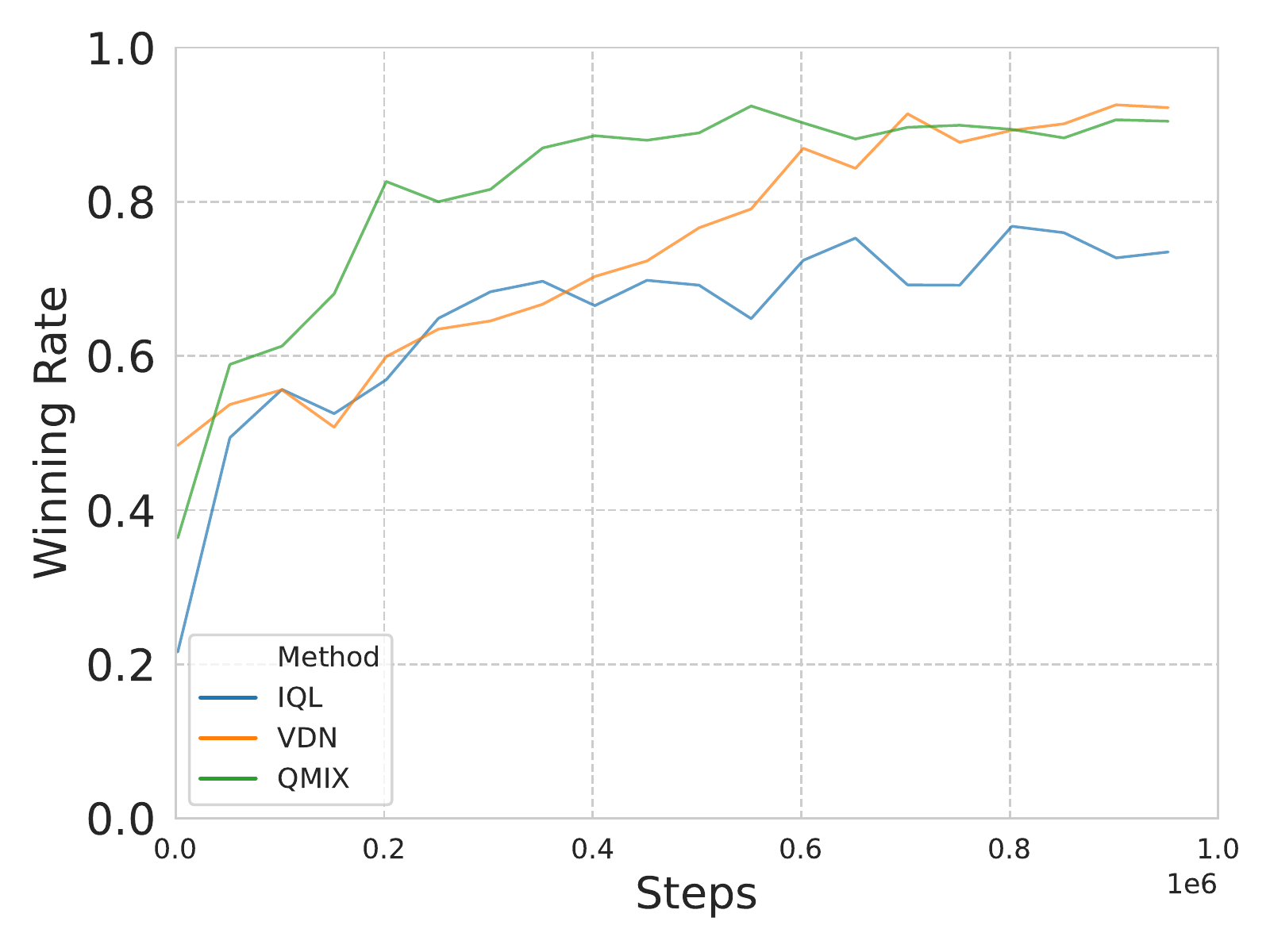}
		}
		\subfigure[{The middle scenario.}]{
			\label{fig:middle}
			\includegraphics[width=2.0in]{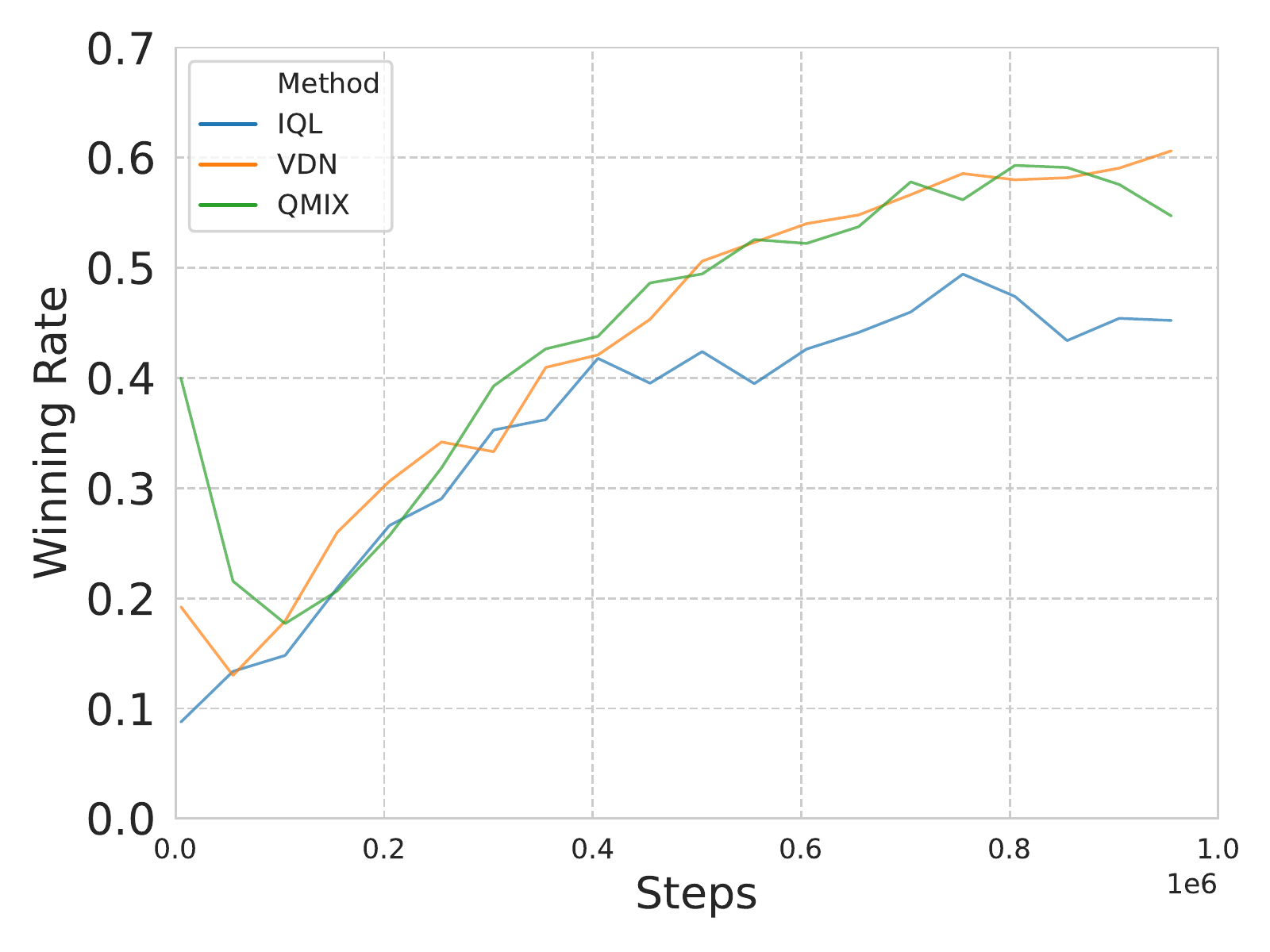}
		}
		\subfigure[{The hard scenario.}]{
			\label{fig:hard}
			\includegraphics[width=2.0in]{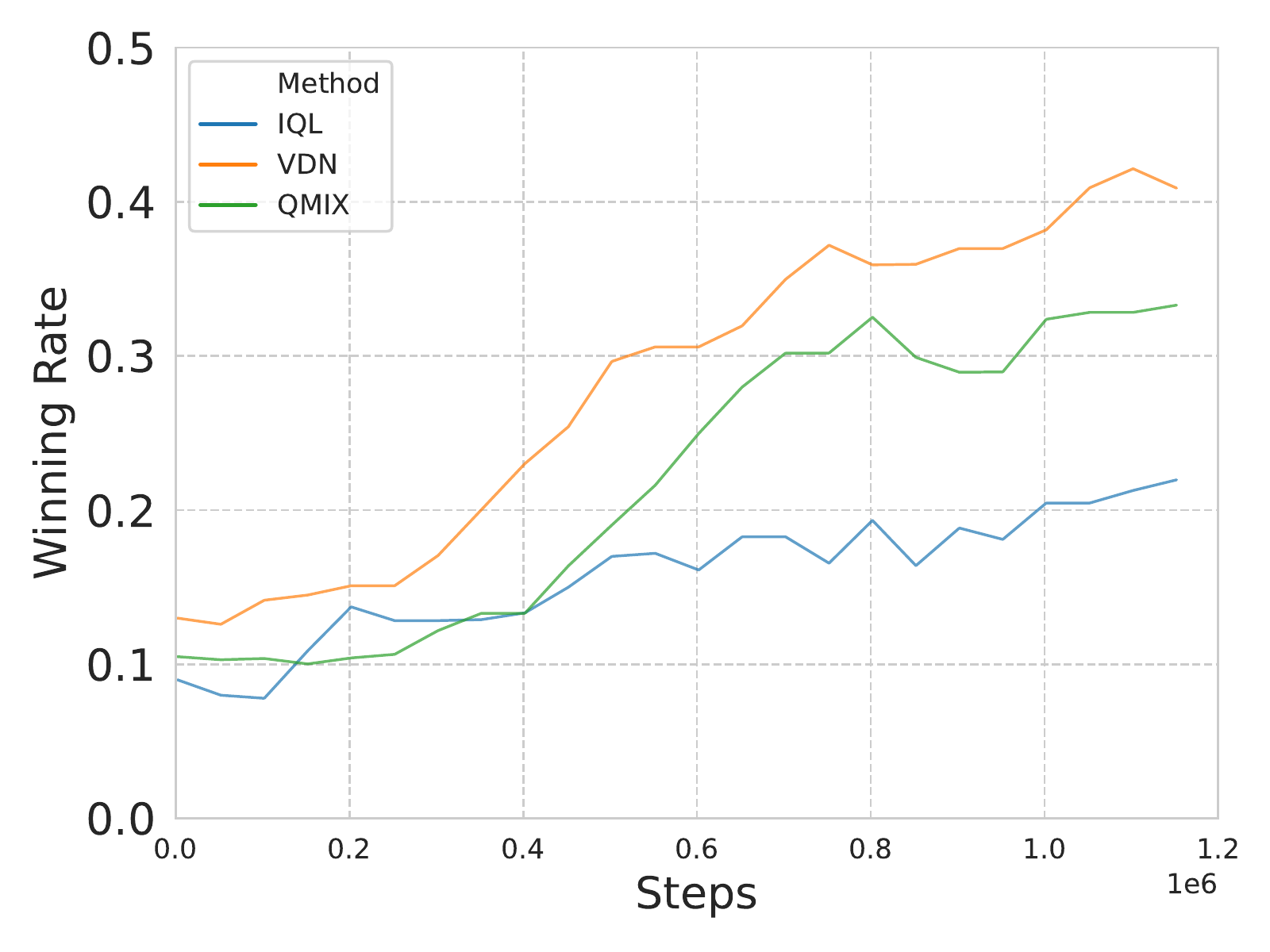}
		}
		\caption{{Learning curves for NeuronsMAE with respect to different algorithms and difficulty levels.} }
		\label{fig:curves}
		
	\end{center}
	\vspace{-0.5cm} 
\end{figure*}

\begin{figure*}[!t]
	\centering
	\begin{center}
		\scriptsize
		\includegraphics[width=6.5in]{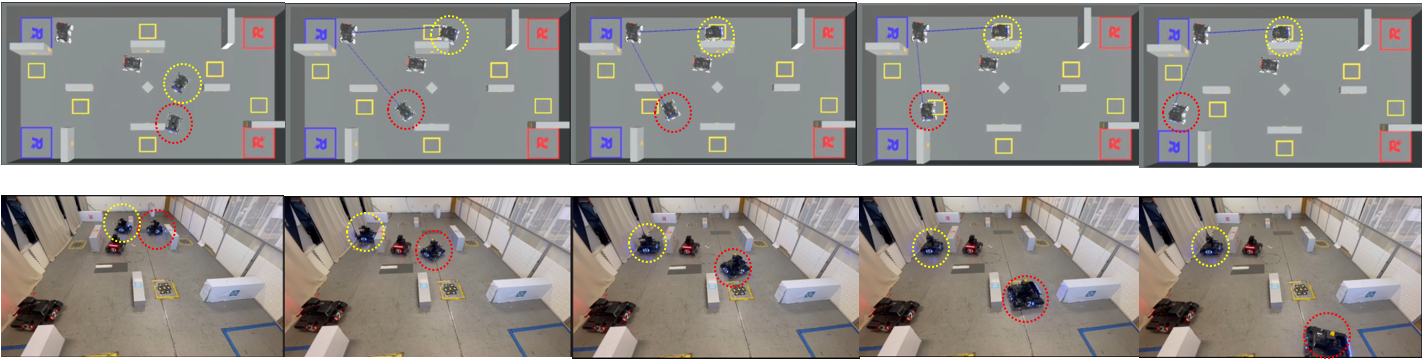}\\
		\caption{Illustration of the trajectories in both simulation and the real world. Two robots besiege an opponent. The top of each subfigure demonstrates how the robots move in the simulation. The bottom of each subfigure shows that the robots cooperate and compete in the real world.}
		\label{fig:s2r}
	\end{center}
	\vspace{-0.5cm} 
\end{figure*}

Figure \ref{fig:curves} shows the learning curves of the five algorithms in three scenarios. It is observed that the trend of performance with the change of the difficulty level manifests itself in a way that is consistent with the fact that the higher the level, the harder it is to defeat the opponent; in addition, this difficulty setting is reasonable.

\subsection{Sim2Real Verification}\label{sec:real}
We expect that the model of all algorithms trained in our simulation can be directly applied in real-world settings. This implementation is commonly referred to as zero-shot or direct transfer and is different from domain randomization, which can be seen as a one-shot transfer. Figure \ref{fig:s2r} shows the poses of robots in several trajectories. Robots move similarly in both simulation and the real world, which reveals that NeuronsMAE is suitable for zero-shot and Sim2Real. In summary, our empirical evaluation reveals that NeuronsMAE is feasible and can enable future developments in MARL.

\section {Further Challenges Enabled by NeuronsMAE}\label{sec:Challenges}
The experiments section only give the tip of the iceberg on how NeuronsMAE can be used to study the cooperation of multi-agent system. NeuronsMAE has various interfaces to choose from, which provides more possibilities for future research. In addition, there will be 5 challenges, and we discuss how NeuronsMAE could be used to tackle them as follows:

\noindent (1) Competitive MARL: It is particularly suitable for training and evaluating competitive MARL methods by replacing the opponent bot with any trained model.

\noindent (2) Continuous action space: NeuronsMAE provides the continuous action space interface, and supports research on continuous control, where actions are continuous and often high-dimensional.

\noindent (3) Sparse reward: it also provides a reward interface for sparse reward research. Sparse reward refers to the problem of agents finding it difficult to obtain positive rewards while exploring, which leads to slow learning or even inability to learn.

\noindent (4) Generalization: The dynamic parameters interface allows users to conduct research into questions, including Sim2Real, Meta RL or continue RL. 

\noindent (5) Multimodal MARL: NeuronsMAE will be helpful in studying multimodal RL with various sensor data. Multimodal RL refers to introducing multimodal data to enhance the performance of RL.

\section {Conclusion}\label{sec:Conclusion}

NeuronsMAE is a research platform designed for studying the cooperative and competitive tasks of robots in the real world, including the highly flexible observation space and action space, and can customize the state observability and action attributes according to the algorithm characteristics. It provides a high-fidelity environment and robot model, as well as rich parameter interfaces, to support the research of multi-robot policy transfer from a simulation environment to a real-world environment. Several commonly used MARL algorithms are evaluated on NeuronsMAE, and a new benchmark for multi-robot cooperative and competitive tasks is established. For future work, we would support more environments with heterogeneous robots and different battlefields and provide more algorithms in NeuronsMAE.

\ifCLASSOPTIONcaptionsoff
\newpage
\fi

\bibliographystyle{IEEEtran}
\small
\bibliography{ref}

\end{document}